\documentclass[10pt,twocolumn,letterpaper]{article}

\usepackage{cvpr}              

\usepackage{graphicx}
\usepackage{amsmath}
\usepackage{amssymb}
\usepackage{booktabs}
\usepackage{multirow}
\usepackage{diagbox}

\usepackage{verbatim}
\usepackage{float}
\usepackage{bm}
\usepackage{xcolor}
\usepackage{enumitem}
\usepackage{verbatim}
\usepackage{pifont}

\usepackage[pagebackref,breaklinks,colorlinks]{hyperref}

\usepackage[pagebackref,breaklinks,colorlinks]{hyperref}

\usepackage[capitalize]{cleveref}
\crefname{section}{Sec.}{Secs.}
\Crefname{section}{Section}{Sections}
\Crefname{table}{Table}{Tables}
\crefname{table}{Tab.}{Tabs.}

\def\cvprPaperID{3094} 
\def\confName{CVPR}
\def\confYear{2023}

\begin{document}

\title{DiffTalk: Crafting Diffusion Models for\\Generalized Audio-Driven Portraits Animation}
\newcommand*\samethanks[1][\value{footnote}]{\footnotemark[#1]}
\author{Shuai Shen$^{1}$\quad Wenliang Zhao$^{1}$\quad Zibin Meng$^1$\quad Wanhua Li${^1}$\quad Zheng Zhu$^2$\quad Jie Zhou$^{1}$\quad Jiwen Lu$^{1,}$$^{\ast}$ \\
$^1$Tsinghua University \quad\quad
$^2$PhiGent Robotics \\}

\twocolumn[{%
\renewcommand\twocolumn[1][]{#1}%
\maketitle
\begin{center}
    \centering
    \captionsetup{type=figure}
    \includegraphics[width=1\textwidth]{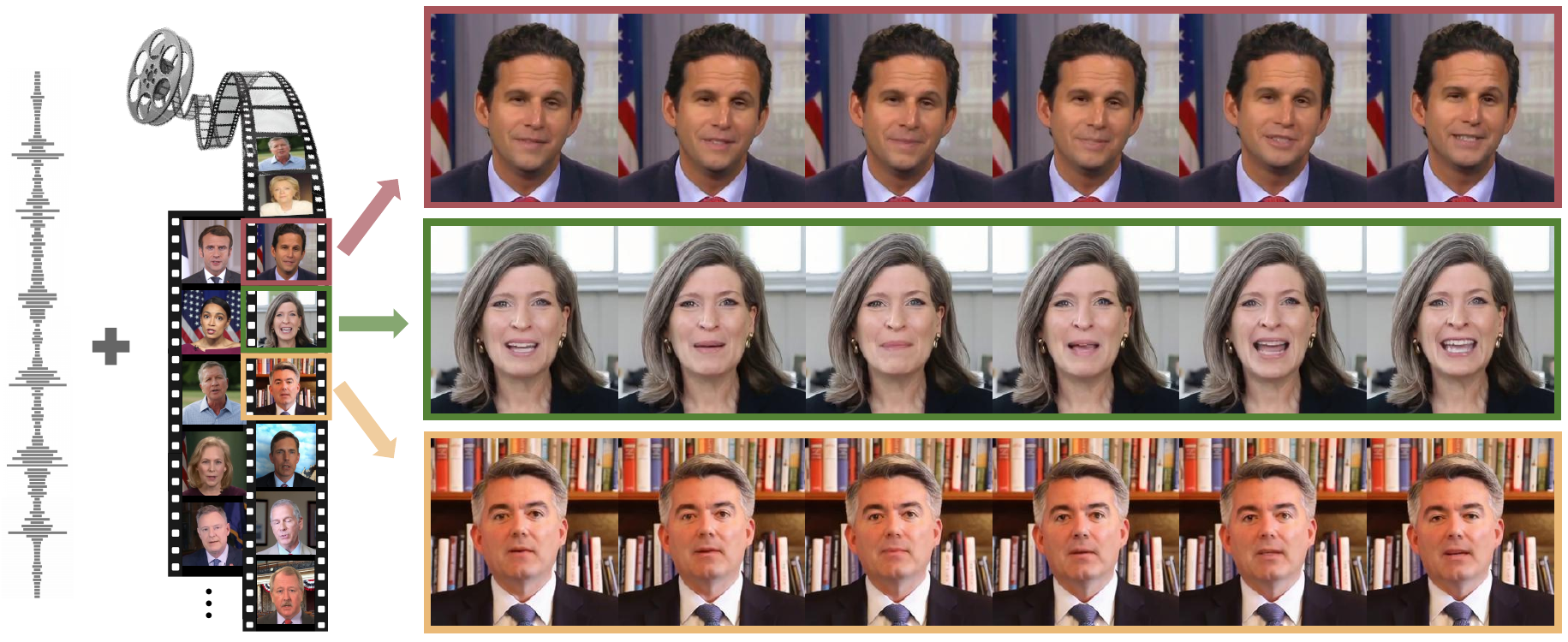}
    \captionof{figure}{We present a crafted conditional \textbf{Diff}usion model for generalized \textbf{Talk}ing head synthesis (DiffTalk). Given a driven audio, the DiffTalk is capable of synthesizing high-fidelity and synchronized talking videos for multiple identities without further fine-tuning.}
    \label{fig1}
\vspace{2mm}
\end{center}%
}]

\let\thefootnote\relax\footnotetext{$^{\ast}$Corresponding author}

\begin{abstract}
Talking head synthesis is a promising approach for the video production industry. Recently, a lot of effort has been devoted in this research area to improve the \textbf{generation quality} or enhance the \textbf{model generalization}. However, there are few works able to address both issues simultaneously, which is essential for practical applications. To this end, in this paper, we turn attention to the emerging powerful Latent \textbf{Diff}usion Models, and model the \textbf{Talk}ing head generation as an audio-driven temporally coherent denoising process (DiffTalk). More specifically, instead of employing audio signals as the single driving factor, we investigate the control mechanism of the talking face, and incorporate reference face images and landmarks as conditions for personality-aware generalized synthesis. In this way, the proposed DiffTalk is capable of producing high-quality talking head videos in synchronization with the source audio, and more importantly, it can be naturally generalized across different identities without further fine-tuning. Additionally, our DiffTalk can be gracefully tailored for higher-resolution synthesis with negligible extra computational cost. Extensive experiments show that the proposed DiffTalk efficiently synthesizes high-fidelity audio-driven talking head videos for generalized novel identities. For more video results, please refer to \url{https://sstzal.github.io/DiffTalk/}.


\end{abstract}

\section{Introduction}
\label{sec:intro}
Talking head synthesis is a challenging and promising research topic, which aims to generate video portraits with given audio. This technique is widely applied in various practical scenarios including animation, virtual avatars, online education, and video conferencing~\cite{zakharov2019few,zhang2020davd,wang2021one,chen2020talking,zhou2020makelttalk}.

Recently a lot of effort has been devoted to this research area to improve the \textbf{generation quality} or enhance the \textbf{model generalization}.
Among these existing mainstream talking head generation approaches, the 2D-based methods usually depend on generative adversarial networks (GANs)~\cite{das2020speech,gu2020flnet,christos2020headgan,prajwal2020lip,kr2019towards} for audio-to-lip mapping, and most of them perform competently on model generalization. However, since GANs need to simultaneously optimize a generator and a discriminator, the training process lacks stability and is prone to mode collapse~\cite{dhariwal2021diffusion}. Due to this restriction, the generated talking videos are of limited image quality, and difficult to scale to higher resolutions.
By contrast, 3D-based methods~\cite{zollhofer2018state,blanz1999morphable,thies2016face2face,guo2021ad,yao2022dfa} perform better in synthesizing higher-quality talking videos. Whereas, they highly rely on identity-specific training, and thus cannot generalize across different persons. Such identity-specific training also brings heavy resource consumption and is not friendly to practical applications. Most recently, there are some 3D-based works~\cite{shen2022dfrf} that take a step towards improving the generalization of the model. However, further fine-tuning on specific identities is still inevitable.

Generation quality and model generalization are two essential factors for better deployment of the talking head synthesis technique to real-world applications. However, few existing works are able to address both issues well. In this paper, we propose a crafted conditional \textbf{Diff}usion model for generalized \textbf{Talk}ing head synthesis (DiffTalk), that aims to tackle these two challenges simultaneously. Specifically, to avoid the unstable training of GANs, we turn attention to the recently developed generative technology Latent Diffusion Models~\cite{rombach2022high}, and model the talking head synthesis as an audio-driven temporally coherent denoising process. On this basis, instead of utilizing audio signals as the single driving factor to learn the audio-to-lip translation, we further incorporate reference face images and landmarks as supplementary conditions to guide the face identity and head pose for personality-aware video synthesis. Under these designs, the talking head generation process is more controllable, which enables the learned model to naturally generalize across different identities without further fine-tuning. As shown in Figure~\ref{fig1}, with a sequence of driven audio, our DiffTalk is capable of producing natural talking videos of different identities based on the corresponding reference videos. Moreover, benefiting from the latent space learning mode, our DiffTalk can be gracefully tailored for higher-resolution synthesis with negligible extra computational cost, which is meaningful for improving the generation quality. 

Extensive experiments show that our DiffTalk can synthesize high-fidelity talking videos for novel identities without any further fine-tuning. Figure~\ref{fig1} shows the generated talking sequences with one driven audio across three different identities.  Comprehensive method comparisons show the superiority of the proposed DiffTalk, which provides a strong baseline for the high-performance talking head synthesis.
To summarize, we make the following contributions:
\begin{itemize}[leftmargin=*]
\setlength{\itemsep}{0pt}
\setlength{\parsep}{0pt}
\setlength{\parskip}{0pt}
\item 
We propose a crafted conditional diffusion model for high-quality and generalized talking head synthesis. By introducing smooth audio signals as a condition, we model the generation as an audio-driven temporally coherent denoising process.

\item For personality-aware generalized synthesis, we further incorporate dual reference images as conditions. In this way, the trained model can be generalized across different identities without further fine-tuning. 

\item The proposed DiffTalk can generate high-fidelity and vivid talking videos for generalized identities. In experiment, our DiffTalk significantly outperforms 2D-based methods in the generated image quality, while surpassing 3D-based works in the model generalization ability.
\end{itemize}

\section{Related Work}

\textbf{Audio-driven Talking Head Synthesis.}
The talking head synthesis aims to generate talking videos with lip movements synchronized with the driving audio~\cite{suwajanakorn2017synthesizing,garrido2015vdub,zhou2021pose}. In terms of the modeling approach, we roughly divide the existing methods into 2D-based and 3D-based ones. In the 2D-based methods, GANs~\cite{prajwal2020lip,das2020speech,gu2020flnet,christos2020headgan} are usually employed as the core technologies for learning the audio-to-lip translation. 
Zhou~\emph{et al.}~\cite{zhou2020makelttalk} introduce a speaker-aware audio encoder for personalized head motion modeling. Prajwal~\emph{et al.}~\cite{prajwal2020lip} boost the lip-visual synchronization with a well-trained Lip-Sync expert~\cite{chung2016out}.
However, since the training process of GANs lacks stability and is prone to mode collapse~\cite{dhariwal2021diffusion}, the generated talking videos are always of limited image quality, and difficult to scale to higher resolutions.
Recently a series of 3D-based methods~\cite{ji2021audio,suwajanakorn2017synthesizing,chen2020talking,thies2020neural,linsen2020ebt} have been developed. \cite{suwajanakorn2017synthesizing,linsen2020ebt,thies2020neural} utilize 3D Morphable Models~\cite{blanz1999morphable} for parametric control of the talking face.
More recently, the emerging Neural radiance fields~\cite{mildenhall2020nerf} provide a new solution for 3D-aware talking head synthesis~\cite{guo2021ad,chan2021pi,liu2022semantic,shen2022dfrf}. However, most of these 3D-based works highly rely on identity-specific training, and thus cannot generalize across different identities. Shen~\emph{et al.}~\cite{shen2022dfrf} have tried to improve the generalization of the model, however, further fine-tuning on specific identities is still inevitable. In this work, we propose a brand-new diffusion model-based framework for high-fidelity and generalized talking head synthesis.

\begin{figure*}[ht]
  \centering
   \includegraphics[width=1\linewidth]{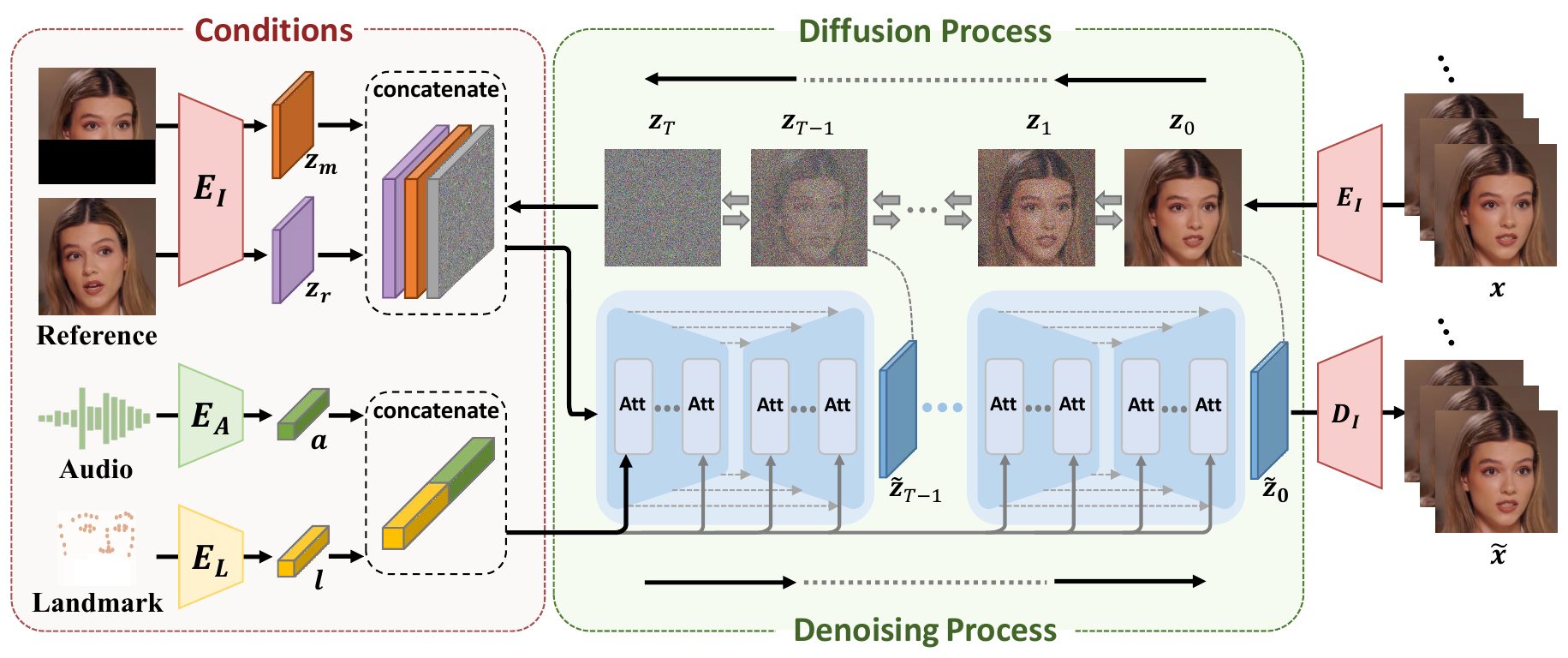}
\vspace{-5mm}
\caption{Overview of the proposed DiffTalk for generalized talking head synthesis. Apart from the audio condition to drive the lip motions, we further incorporate reference images and facial landmarks as extra driving factors for personalized facial modeling. In this way, the generation process is more controllable, which enables the learned model to generalize across different identities without further fine-tuning. Furthermore, we can gracefully improve our DiffTalk for higher-resolution synthesis with slight extra computational cost.}
\vspace{-2mm}
\label{fig2}
\end{figure*}

\textbf{Latent Diffusion Models.}
Diffusion Probabilistic Models (DM)~\cite{sohl2015deep} have shown strong ability in various image generation tasks~\cite{dhariwal2021diffusion,ho2020denoising,ramesh2021zero}. However, due to pixel space-based training~\cite{ronneberger2015u,rombach2022high}, very high computational costs are inevitable. More recently, Rombach~\emph{et al.}~\cite{rombach2022high} propose the Latent Diffusion Models (LDMs), and transfer the training and inference processes of DM to a compressed lower-dimension latent space for more efficient computing~\cite{esser2021taming,zhang2018unreasonable}. With the democratizing of this technology, it has been successfully employed in a series of works, including text-to-image translation~\cite{ruiz2022dreambooth,kawar2022imagic,rombach2022text}, super resolution~\cite{dos2022face,pandey2022diffusevae,chung2022diffusion}, image inpainting~\cite{lugmayr2022repaint,ku2022intelligent}, motion generation~\cite{zhang2022motiondiffuse,shao2022diffustereo}, 3D-aware prediction~\cite{waibel2022diffusion,bautista2022gaudi,saadatnejad2022generic}. In this work, drawing on these successful practices, we model the talking head synthesis as an audio-driven temporally coherent denoising process and achieve superior generation results.


\section{Methodology}


\subsection{Overview}
\label{overview}

To tackle the challenges of generation quality and model generalization, we model the talking head synthesis as an audio-driven temporally coherent denoising process, and term the proposed method as DiffTalk. An overview of the DiffTalk is shown in Figure~\ref{fig2}. By introducing smooth audio features as a condition, we improve the diffusion model for temporally coherent facial motion modeling. For further personalized facial modeling, we incorporate reference face images and facial landmarks as extra driving factors. In this way, the talking head generation process is more controllable, which enables the learned model to generalize across different identities without any further fine-tuning. Moreover, benefiting from the latent space learning mode, we can graceful improve our DiffTalk for higher-resolution synthesis with negligible extra computational cost, which contributes to improving the generation quality.

\subsection{Conditional Diffusion Model for Talking Head}
\label{condition}

The emergence of Latent Diffusion Models (LDMs)~\cite{ho2020denoising,rombach2022high}
provides a straightforward and effective way for high-fidelity image synthesis. To inherit its excellent properties, we adopt this advanced technology as the foundation of our method and explore its potential in modeling the dynamic talking head. With a pair of well-trained image encoder $E_I$ and decoder $D_I$ which are frozen in training~\cite{esser2021taming}, the input face image $x \in \mathbb{R}^{H\times W\times 3}$ can be encoded into a latent space $z_0 = E_I (x) \in \mathbb{R}^{h\times w\times 3}$, where $H/h = W/w = f$, $H,W$ are the height and width of the original image and $f$ is the downsampling factor. In this way, the learning is transferred to a lower-dimensional latent space, which is more efficient with fewer train resources. On this basis, the standard LDMs are modeled as a time-conditional UNet-based~\cite{ronneberger2015u} denoising network $\mathcal{M}$, which learns the reverse process of a Markov Chain~\cite{geyer1992practical} of length $T$. The corresponding objective can be formulated as:
\begin{equation}
L_{L D M}:=\mathbb{E}_{z, \epsilon \sim \mathcal{N}(0,1), t}\left[\left\|\epsilon-\mathcal{M}\left(z_{t}, t\right)\right\|_{2}^{2}\right],
\end{equation}
where $t\in [1, \cdots, T]$ and $z_t$ is obtained through the forward diffusion process from $z_0$. $\tilde{z}_{t-1}=z_t - \mathcal{M} (z_{t}, t)$ is the denoising result of $z_t$ at time step $t$. The final denoised result $\tilde{z}_0$ is then upsampled to the pixel space with the pre-trained image decoder $\tilde{x} = D_I (\tilde{z}_0)$, where $\tilde{x}\in \mathbb{R}^{H\times W\times 3}$ is the reconstructed face image.

Given a source identity and driven audio, our goal is to train a model for generating a natural target talking video in synchronization with the audio condition while maintaining the original identity information. Furthermore, the trained model also needs to work for novel identities during inference. To this end, the audio signal is introduced as a basic condition to guide the direction of the denoising process for modeling the audio-to-lip translation. 

\begin{figure}[t]
  \centering
   \includegraphics[width=1\linewidth]{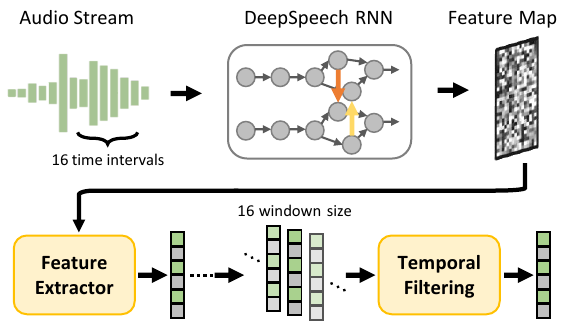}
   \caption{Visualization of the smooth audio feature extractor.}
   \label{fig3}
   \vspace{-2mm}
\end{figure}

\vspace{1mm}
\textbf{Smooth Audio Feature Extraction.}
To better incorporate temporal information, we involve two-stage smoothing operations in the audio encoder $E_A$, as shown in Figure~\ref{fig3}. Firstly, following the practice in VOCA~\cite{cudeiro2019capture}, we reorganize the raw audio signal into overlapped windows of size 16 time intervals (corresponding to audio clips of 20ms), where each window is centered on the corresponding video frame. A pre-trained RNN-based DeepSpeech~\cite{hannun2014deep} module is then leveraged to extract the per-frame audio feature map $F$. For better inter-frame consistency, we further introduce a learnable temporal filtering~\cite{thies2020neural}. It receives a sequence of adjacent audio features $[F_{i-w}, \dots, F_i, \dots, F_{i+w}]$ with $w=8$ as input, and computes the final smoothed audio feature for the $i$-th frame as $a \in \mathbb{R}^{D_A}$ in a self-attention-based learning manner, where $D_A$ denotes the audio feature dimension.
By encoding the audio information, we bridge the modality gap between the audio signals and the visual information. Introducing such smooth audio features as a condition, we extend the diffusion model for temporal coherence-aware modeling of face dynamics when talking. The objective is then formulated as:
\begin{equation}
L_A:=\mathbb{E}_{z, \epsilon \sim \mathcal{N}(0,1), a, t}\left[\left\|\epsilon-\mathcal{M}\left(z_{t}, t, a\right)\right\|_{2}^{2}\right].
\label{eq:important}
\end{equation}

\vspace{1mm}
\textbf{Identity-Preserving Model Generalization.}
In addition to learning the audio-to-lip translation, another essential task is to realize the model generalization while preserving complete identity information in the source image. Generalized identity information includes face appearance, head pose, and image background. To this end, a reference mechanism is designed to empower our model to generalize to new individuals unseen in training, as shown in Figure~\ref{fig2}. Specifically, a random face image $x_r$ of the source identity is chosen as a reference condition, which contains appearance and background information. To prevent training shortcuts, we limit the selection of $x_r$ to 60 frames beyond the target image. However, since the ground-truth face image has a completely different pose from $x_r$, the model is expected to transfer the pose of $x_r$ to the target face without any prior information. This is somehow an ill-posed problem with no unique solution. For this reason, we further incorporate the masked ground-truth image $x_m$ as another reference condition to provide the target head pose guidance. The mouth region of $x_m$ is completely masked to ensure that the ground truth lip movements are not visible to the network. In this way, the reference $x_r$ focuses on affording mouth appearance information, which additionally reduces the training difficulty. Before serving as conditions, $x_r$ and $x_m$ are also encoded into the latent space through the trained image encoder, and we have $z_r = D_I (x_r)\in \mathbb{R}^{h\times w\times 3}$, $z_m = D_I (x_m)\in \mathbb{R}^{h\times w\times 3}$.
On this basis, an auxiliary facial landmarks condition is also included for better control of the face outline. Similarly, landmarks in the mouth area are masked to avoid shortcuts. The landmark feature $l\in \mathbb{R}^{D_L}$ is obtained with an MLP-based encoder $E_L$, where $D_L$ is the landmark feature dimension. In this way, combining these conditions with audio feature $a$, we realize the precise control over all key elements of a dynamic talking face. With $C=\{a, z_r, z_m, l\}$ denoting the condition set, the talking head synthesis is finally modeled as a conditional denoising process optimized with the following objective: 
\begin{equation}
L:=\mathbb{E}_{z, \epsilon \sim \mathcal{N}(0,1), C, t}\left[\left\|\epsilon-\mathcal{M}\left(z_{t}, t, C\right)\right\|_{2}^{2}\right],
\label{final_eq}
\end{equation}
where the network parameters of $\mathcal{M}, E_A$ and $E_L$ are jointly optimized via this equation.

\vspace{1mm}
\textbf{Conditioning Mechanisms.}
Based on the modeling of the conditional denoising process in Eq.~\ref{final_eq}, we pass these conditions $C$ to the network in the manner shown in Figure~\ref{fig2}. Specifically, following~\cite{rombach2022high}, we implement the UNet-based backbone $\mathcal{M}$ with the cross-attention mechanism for better multimodality learning. The spatially aligned references $z_r$ and $z_m$ are concatenated channel-wise with the noisy map $z_T$ to produce a joint visual condition $C_v = [z_T; z_m; z_r] \in \mathbb{R}^ {h\times w\times 9}$. $C_v$ is fed to the first layer of the network to directly guide the output face in an image-to-image translation fashion. Additionally, the driven-audio feature $a$ and the landmark representation $l$ are concatenated into a latent condition $C_l=[a; l] \in \mathbb{R}^{D_A + D_L}$, which serves as the \emph{key} and \emph{value} for the intermediate cross-attention layers of $\mathcal{M}$. To this extent, all condition information $C=\{C_v, C_l\}$ are properly integrated into the denoising network $\mathcal{M}$ to guide the talking head generation process.


\begin{figure}[t]
  \centering
   \includegraphics[width=1\linewidth]{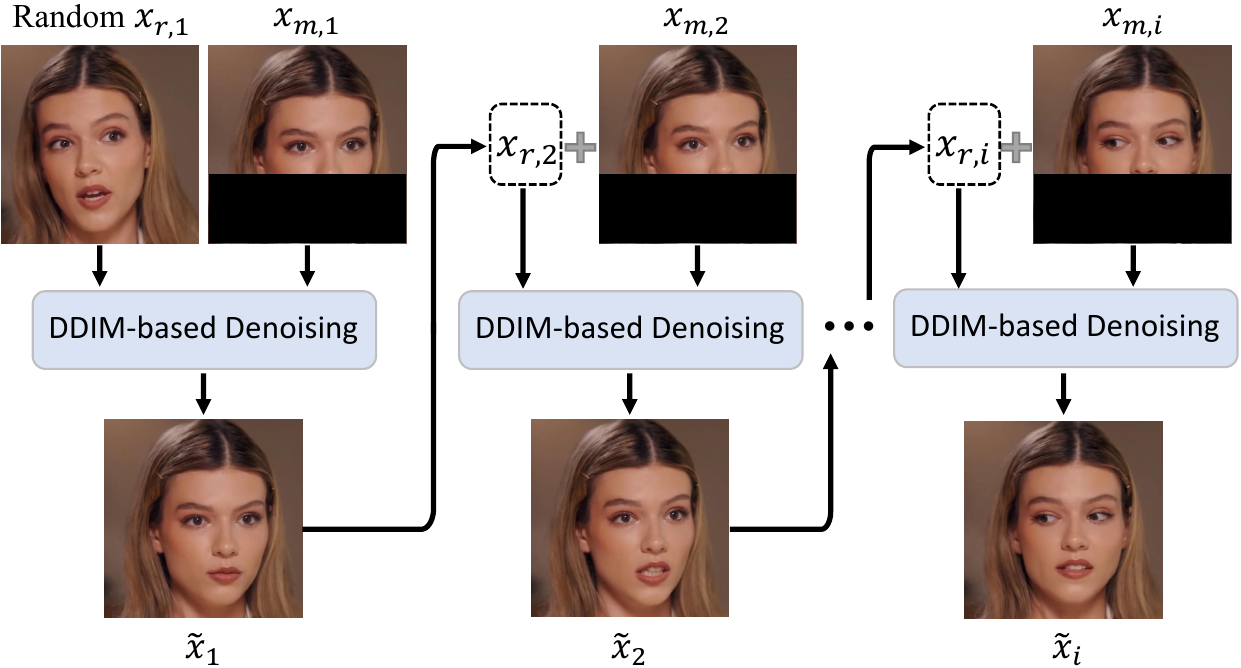}
   \vspace{-4mm}
   \caption{Illustration of the progressive inference strategy.}
   \label{fig4}
   \vspace{-3mm}
\end{figure}

\vspace{1mm}
\textbf{Higher-Resolution Talking Head Synthesis}
\label{details}
Our proposed DiffTalk can also be gracefully extended for higher-resolution talking head synthesis with negligible extra computational cost and faithful reconstruction effects. Specifically, considering the trade-off between the perceptual loss and the compression rate, for training images of size $256\times 256\times 3$, we set the downsampling factor as $f=4$ and obtain the latent space of $64\times 64 \times 3$. Furthermore, for higher-resolution generation of $512\times 512\times 3$, we just need to adjust the paired image encoder $E_I$ and decoder $D_I$ with a bigger downsampling factor $f=8$. Then the trained encoder is frozen and employed to transfer the training process to a $64\times 64 \times 3$ latent space as well. This helps to relieve the pressure on insufficient resources, and therefore our model can be gracefully improved for higher-resolution talking head video synthesis.




\subsection{Progressive Inference}
\label{reference}
We perform inference with Denoising Diffusion Implicit Model-based (DDIM)~\cite{song2020denoising} iterative denoising steps to accelerate sampling for more efficient synthesis. 
To further boost the coherence of the generated talking videos, we develop a progressive reference strategy in the reference process as shown in Figure~\ref{fig4}. Specifically, when rendering a talking video sequence with the trained model, for the first frame, $x_{r,1}$ is a random face image from the target identity. Subsequently, the synthetic face image $\tilde{x}_{i}$ is exploited as the reference $x_{r,i+1}$ for the next frame. In this way, image details between adjacent frames remain consistent, resulting in a smoother transition between frames. It is worth noting that this strategy is not used for training. Since the difference between adjacent frames is small, we need to eliminate such references to avoid learning shortcuts. Following the practice in~\cite{rombach2022high}, masked $z_T$ is used during inference, where the mouth area is masked and randomly initialized, allowing the network to focus on the denoising of this region. To further alleviate the video jitter issue, we utilize~\cite{huang2022rife} for frame interpolation to get smoother synthesized talking videos.

\begin{figure}[t]
  \centering
   \includegraphics[width=1\linewidth]{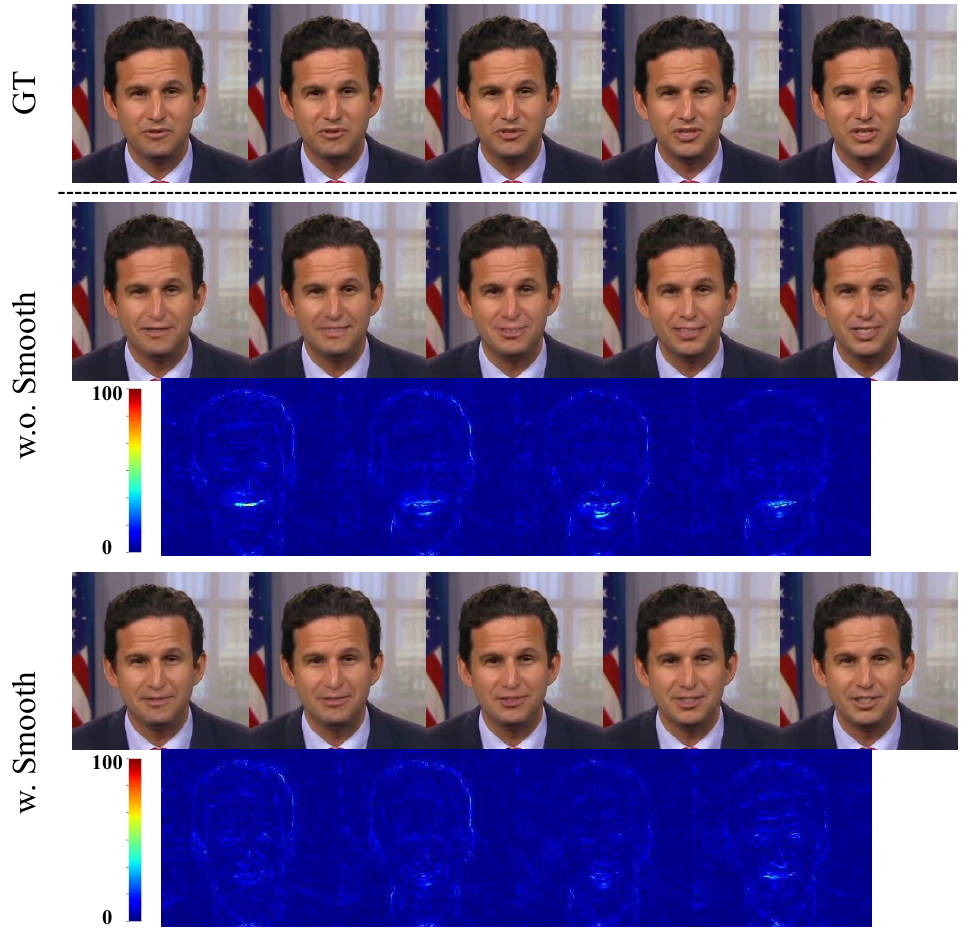}
   \vspace{-6mm}
   \caption{Ablation study on the audio smoothing operation. We show the differences between adjacent frames as heatmaps for better visualization.}
   \label{fig5}
\end{figure}

\section{Experiments}

\subsection{Experimental Settings}
\textbf{Dataset. }To train the audio-driven diffusion model, an audio-visual dataset HDTF~\cite{zhang2021flow} is used. It contains 16 hours of talking videos in 720P or 1080P from more than 300 identities. We randomly select 100 videos, and finally form a video gallery with the length of 100 minutes for training, while the remaining data serve as the test set.

\begin{figure*}[t]
  \centering
   \includegraphics[width=1\linewidth]{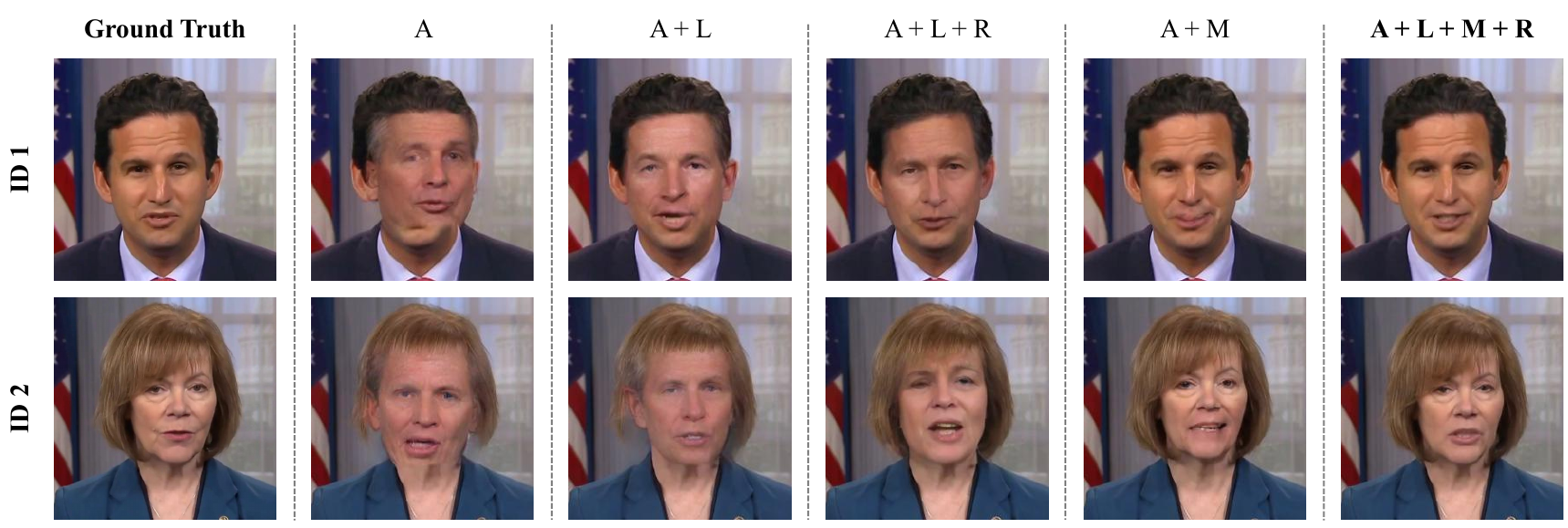}
   \vspace{-6mm}
   \caption{Ablation study on the design of the conditions. The marks above these images refer to the following meanings, `A': Audio; `L': Landmark; `R': Random reference image; `M': Masked ground-truth image. We show the generated results under different condition settings on two test sets, and demonstrate the effectiveness of our final design, \ie A+L+M+R.}
   \label{fig6}
\end{figure*}

\textbf{Metric.}
We evaluate our proposed method through visual results coupled with quantitative indicators. PSNR ($\uparrow$), SSIM ($\uparrow$)~\cite{wang2004image} and LPIPS ($\downarrow$)~\cite{zhang2018unreasonable} are three metrics for assessing image quality. The LPIPS is a learning-based perceptual similarity measure that is more in line with human perception, we therefore recommend this metric as a more objective indicator. The SyncNet score (Offset$\downarrow$ / Confidence$\uparrow$)~\cite{chung2016out} checks the audio-visual synchronization quality, which is  important for the audio-driven talking head generation task. (`$\downarrow$' indicates that the lower
the better, while `$\uparrow$' means that the higher the better.)

\begin{table}[tb]
\begin{center}
\centering
\renewcommand\tabcolsep{3.8pt}
\begin{tabular}{c |c|c c c c} 
\toprule
 \multicolumn{2}{c|}{Method}&PSNR$\uparrow$ &SSIM$\uparrow$ & LPIPS$\downarrow$  & SyncNet$\downarrow\uparrow$  \\
\midrule
\multirow{3}{*}{Test Set A}&GT&-&-&-&0/9.610\\
&w/o& 33.67&0.944 &\bf{0.024} &1/5.484\\
&w&\bf{34.17}&\bf{0.946}&\bf{0.024}&\bf{1/6.287}\\
\midrule
\multirow{3}{*}{Test Set B}&GT&-&-&-&0/9.553\\
&w/o & 32.70&\bf{0.924}&\bf{0.031}&1/5.197 \\
&w&\bf{32.73}&\bf{0.925}&\bf{0.031}&\bf{1/5.387}\\
\bottomrule
\end{tabular}
\end{center}
\vspace{-5mm}
\caption{Ablation study to investigate the contribution of the audio smoothing operation. ‘w’ indicates the model is trained with the audio features after temporal filtering and vice versa.}
\vspace{-3mm}
\label{table1}
\end{table}

\textbf{Implementation Details.}
We resize the input image to $256\times 256$ for experiments. The downsampling factor $f$ is set as $4$, so the latent space is $64\times 64 \times 3$. For training the model for higher resolution synthesis, the input is resized to $512\times 512$ with $f=8$ to keep the same size of latent space.
The length of the denoising step $T$ is set as 200 for both the training and inference process. The feature dimensions are $D_A=D_L=64$.
Our model takes about 15 hours to train on 8 NVIDIA 3090 GPUs. 

\subsection{Ablation Study}

\textbf{Effect of the Smooth Audio.} In this subsection, we investigate the effect of the audio smooth operations. Quantitative results in Table~\ref{table1} show that the model equipped with the audio temporal filtering module outperforms the one without smooth audio, especially in the SyncNet score. We further visualize the differences between adjacent frames as the heatmaps shown in Figure~\ref{fig5}. The results without audio filtering present obvious high heat values in the mouth region, which indicates the jitters in this area. By contrast, with smooth audio as the condition, the generated video frames show smoother transitions, which are reflected in the soft differences of adjacent frames.

\begin{table}[tb]
\begin{center}
\centering
\renewcommand\tabcolsep{3.8pt}
\begin{tabular}{c |c|c c c c} 
\toprule
 \multicolumn{2}{c|}{Method}&PSNR$\uparrow$ &SSIM$\uparrow$ & LPIPS$\downarrow$  & SyncNet$\downarrow\uparrow$  \\
\midrule
\multirow{3}{*}{Test Set A}&GT&-&-&-&4/7.762\\
&w/o& \bf{34.17}&\bf{0.946}&0.024&1/6.287\\
&w&33.95&\bf{0.946}&\bf{0.023}&\bf{-1/6.662}\\
\midrule
\multirow{3}{*}{Test Set B}&GT&-&-&-&3/8.947\\
&w/o & 32.73&\bf{0.925}&0.031&1/5.387\\
&w&\bf{33.02}&\bf{0.925}&\bf{0.030}&\bf{1/5.999}\\
\bottomrule
\end{tabular}
\end{center}
\vspace{-5mm}
\caption{Ablation study on the effect of the progressive inference strategy. `w/o' indicates that a random reference image is employed as the condition, and `w' means that the reference is the generated result of the previous frame.}
\vspace{-3mm}
\label{table2}
\end{table}

\begin{figure*}[t]
  \centering
   \includegraphics[width=1\linewidth]{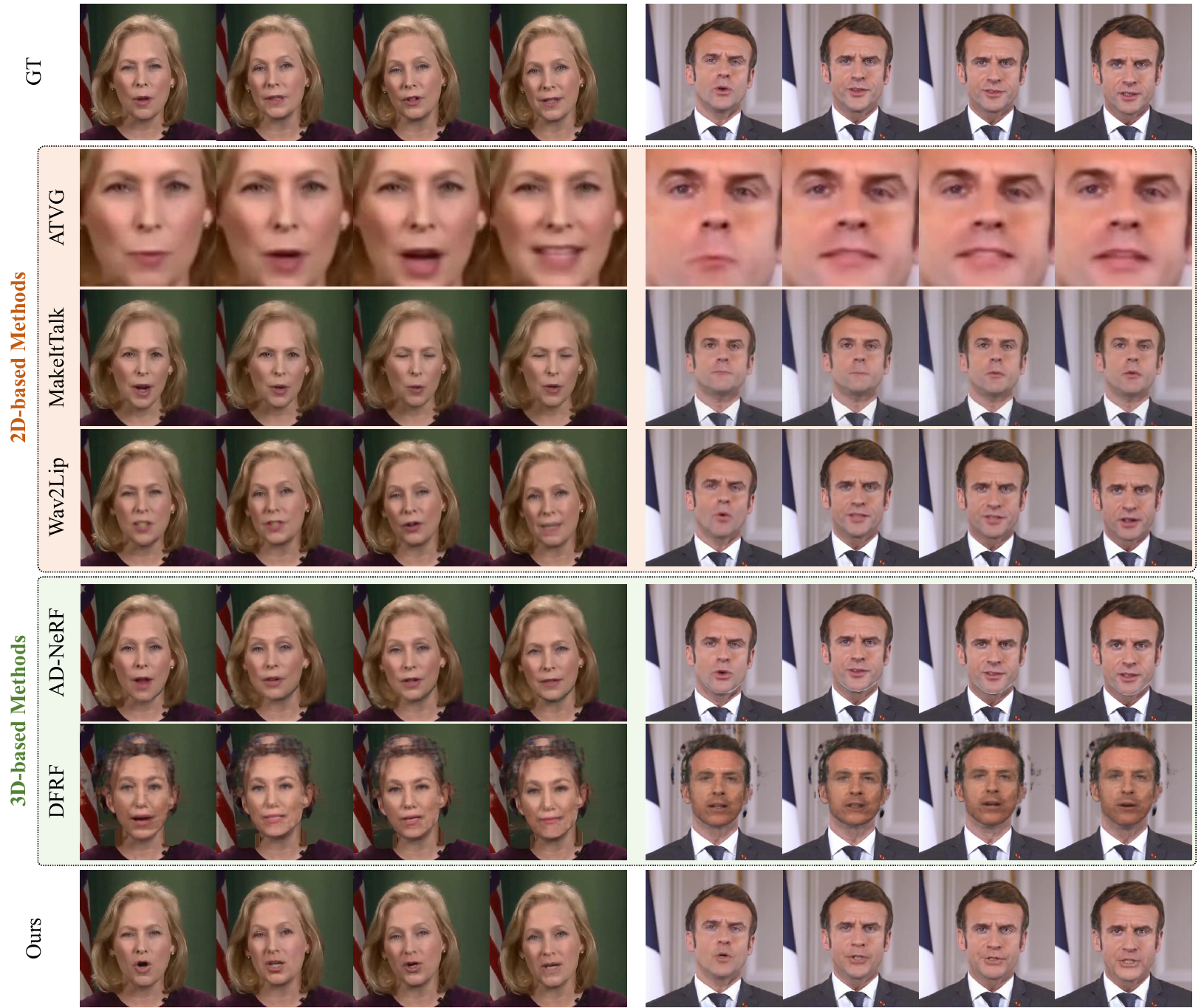}
   \vspace{-6mm}
   \caption{Visual comparison with some representative 2D-based talking head generation methods ATVGnet~\cite{chen2019hierarchical}, MakeitTalk~\cite{zhou2020makelttalk} and Wav2Lip~\cite{prajwal2020lip}, and with some recent 3D-based ones AD-NeRF~\cite{guo2021ad} and DFRF~\cite{shen2022dfrf}. The results of DFRF are synthesized with the base model without fine-tuning for fair comparisons. AD-NeRF is trained on these two identities respectively to produce the results.}
\vspace{-4mm}
   \label{fig7}
\end{figure*}

\textbf{Design of the Conditions.}
A major contribution of this work is the ingenious design of the conditions for general and high-fidelity talking head synthesis. In Figure~\ref{fig6}, we show the generated results under different condition settings step by step, to demonstrate the superiority of our design. With pure audio as the condition, the model fails to generalize to new identities, and the faces are not aligned with the background in the inpainting-based inference. Adding landmarks as another condition tackles the misalignment problem. A random reference image is further introduced trying to provide the identity information. Whereas, since the ground-truth face image has a different pose from this random reference, the model is expected to transfer the pose of reference to the target face. This greatly increases the difficulty of training, leading to hard network convergence, and the identity information is not well learned. Using the audio and masked ground-truth images as driving factors mitigates the identity inconsistency and misalignment issues, however the appearance of the mouth can not be learned since this information is not visible to the network. For this reason, we employ the random reference face and the masked ground-truth image together for dual driving, where the random reference provides the lip appearance message and the masked ground-truth controls the head pose and identity. Facial landmarks are also incorporated as a condition that helps to model the facial contour better. Results in Figure~\ref{fig6} show the effectiveness of such design in synthesizing realism and controllable face images.

\textbf{Impact of the Progressive Inference.}
Temporal correlation inference is developed in this work through the progressive reference strategy. We conduct an ablation study in  Table~\ref{table2} to investigate the impact of this design. `w/o' indicates that a random reference image $x_r$ is employed, and `w' means that the generated result of the previous frame is chosen as the reference condition. With such progressive inference, the SyncNet scores are further boosted, 
since the temporal correlation is better modeled and the talking style becomes more coherent. The LPIPS indicator is also enhanced with this improvement. PSNR tends to give higher scores to blurry images~\cite{zhang2018unreasonable},
so we recommend LPIPS as a more representative metric for visual quality.

\subsection{Method Comparison}
\begin{table*}[t]
\begin{center}
\centering
\begin{tabular}{c|c c c c |c c c c |c} 
\toprule
\multirow{2}{*}{Method} & \multicolumn{4}{c|}{Test Set A}  &\multicolumn{4}{c|}{Test Set B} & {General}\\
 &PSNR$\uparrow$ &SSIM$\uparrow$ & LPIPS$\downarrow$& SyncNet$\downarrow\uparrow$  & PSNR$\uparrow$ &SSIM$\uparrow$ & LPIPS$\downarrow$& SyncNet$\downarrow\uparrow$ &Method \\
\midrule
GT &-&-&-&-1/8.979& -&- &-& -2/7.924&-\\
MakeItTalk~\cite{zhou2020makelttalk}&18.77&0.544&0.19&-4/3.936&17.70&0.648&0.129&-3/3.416&\checkmark\\
Wav2Lip~\cite{prajwal2020lip}&25.50&0.761&0.140&\bf{\textcolor{blue}{-2}}/\bf{\textcolor{red}{8.936}}&33.38&0.942&0.027&-3/\bf{\textcolor{red}{9.385}}&\checkmark\\
AD-NeRF~\cite{guo2021ad}&27.89&0.885&0.072&\textbf{\textcolor{blue}{-2}}/{5.639}&30.14&0.947&\bf{\textcolor{blue}{0.023}}&-3/4.246&\ding{53}\\
DFRF~\cite{shen2022dfrf}&\bf{\textcolor{blue}{28.60}}&\bf{\textcolor{blue}{0.892}}&\bf{\textcolor{blue}{0.068}}&\textbf{\textcolor{red}{-1}}/5.999&\bf{\textcolor{blue}{33.57}}&\bf{\textcolor{blue}{0.949}}&0.025&\textbf{\textcolor{blue}{-2}}/4.432& FT Req.\\
\textbf{Ours}&
\bf{\textcolor{red}{34.54}}&\bf{\textcolor{red}{0.950}}&\bf{\textcolor{red}{0.024}}&\bf{\textcolor{red}{-1}}/\bf{\textcolor{blue}{6.381}}&\bf{\textcolor{red}{34.01}}&\bf{\textcolor{red}{0.950}}&\bf{\textcolor{red}{0.020}}&\bf{\textcolor{red}{-1}}/\bf{\textcolor{blue}{5.639}}&\checkmark\\
\bottomrule
\end{tabular}
\end{center}
\vspace{-4mm}
\caption{
Comparison with some representative talking head synthesis methods on two test sets as in Figure~\ref{fig7}. The best performance is \textbf{highlighted} in \textbf{\textcolor{red}{red}} (1st best) and \textbf{\textcolor{blue}{blue}} (2nd best). Our DiffTalk obtains the best PSNR, SSIM, and LPIPS values, and comparable SyncNet scores simultaneously. It is worth noting that the DFRF is fine-tuned on the specific identity to obtain these results, while our method can directly be utilized for generation without further fine-tuning. (`FT Req.' means that fine-tuning operation is required for the DFRF.)}
\vspace{-2mm}
\label{table3}
\end{table*}

\textbf{Comparison with 2D-based Methods.} In this section, we perform method comparisons with some representative 2D-based talking head generation approaches including the ATVGnet~\cite{chen2019hierarchical}, MakeitTalk~\cite{zhou2020makelttalk} and Wav2Lip~\cite{prajwal2020lip}. Figure~\ref{fig7} visualizes the generated frames of these methods. It can be seen that the ATVGnet performs generation based on cropped faces with limited image quality. The MakeItTalk synthesizes plausible talking frames, however the background is wrongly wrapped with the mouth movements, which greatly affects the visual experience. Generated talking faces of Wav2Lip appear artifacts in the square boundary centered on the mouth, since the synthesized area and the original image are not well blended. By contrast, the proposed DiffTalk generates natural and realistic talking videos with accurate audio-lip synchronization, owing to the crafted conditioning mechanism and stable training process. For more objective comparisons, we further evaluate the quantitative results in Table~\ref{table3}. Our DiffTalk far surpasses \cite{prajwal2020lip} and \cite{zhou2020makelttalk} in all image quality metrics. For the audio-visual synchronization metric SyncNet, the proposed method reaches a high level and is superior to MakeItTalk. Although the DiffTalk is slightly inferior to Wav2Lip on SyncNet score, it is far better than Wav2Lip in terms of image quality. In conclusion, our method outperforms these 2D-based methods under comprehensive consideration of the qualitative and quantitative results.

\textbf{Comparison with 3D-based Methods.}  For more comprehensive
evaluations, we further compare with some recent high-performance 3D-based works including AD-NeRF~\cite{guo2021ad} and DFRF~\cite{shen2022dfrf}. They realize implicitly 3D head modeling with the NeRF technology, so we treat them as generalized 3D-based methods. The visualization results are shown in Figure~\ref{fig7}. AD-NeRF models the head and torso parts separately, resulting in misalignment in the neck region. More importantly, it is worth noting that AD-NeRF is a non-general method. In contrast, our method is able to handle unseen identities without further fine-tuning, which is more in line with the practical application scenarios. The DFRF relies heavily on the fine-tuning operation for model generalization, and the generated talking faces with only the base model are far from satisfactory as shown in Figure~\ref{fig7}. More quantitative results in Table~\ref{table3} also show that our method surpasses \cite{guo2021ad,shen2022dfrf} on the image quality and audio-visual synchronization indicators.

\begin{figure}[t]
  \centering
   \includegraphics[width=1\linewidth]{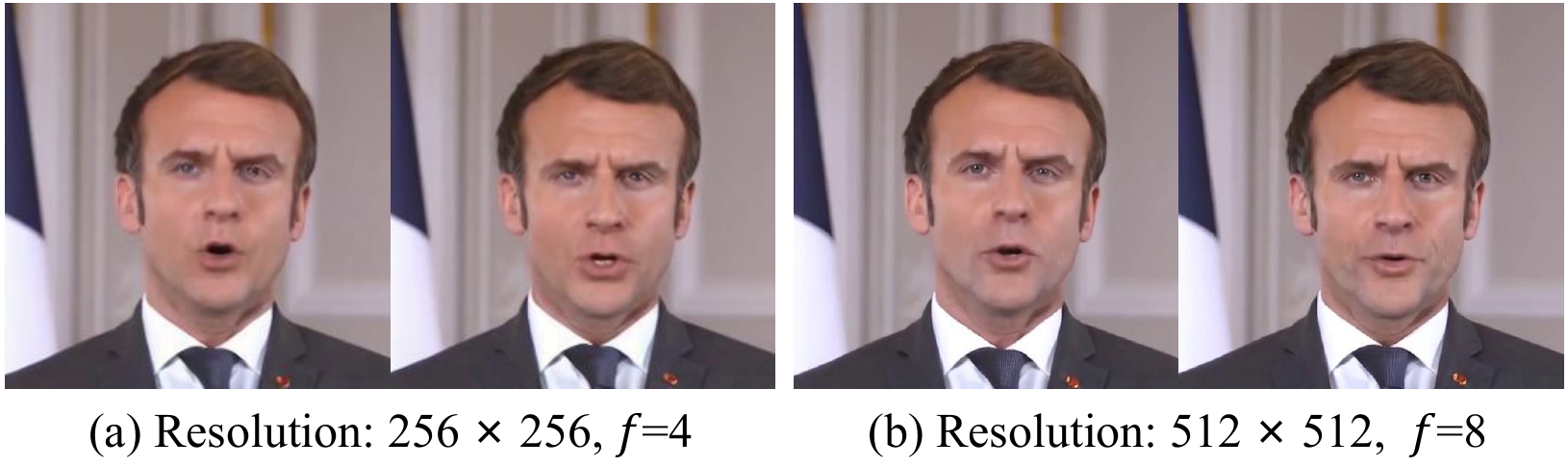}
   \vspace{-6mm}
   \caption{Generated results with higher resolution.}
   \vspace{-4mm}
   \label{fig8}
\end{figure}

\subsection{Expand to Higher Resolution}
In this section, we perform experiments to demonstrate the capacity of our method on generating higher-resolution images. In Figure~\ref{fig8}, we show the synthesis frames of two models (a) and (b). (a) is trained on $256\times 256$ images with the downsampling factor $f=4$, so the latent space is of size $64\times 64\times 3$. For (b), $512\times 512$ images with $f=8$ are used for training the model. Since both models are trained based on a compressed $64\times 63 \times 3$ latent space, the pressure of insufficient computing resources is relieved. We can therefore comfortably expand our model for higher-resolution generation just as shown in Figure~\ref{fig8}, where the synthesis quality in (b) significantly outperforms that in (a).


\section{Conclusion and Discussion}
In this paper, we have proposed a generalized and high-fidelity talking head synthesis method based on a crafted conditional diffusion model. 
Apart from the audio signal condition to drive the lip motions, we further incorporate reference images as driving factors to model the personalized appearance, which enables the learned model to comfortably generalize across different identities without any further fine-tuning. Furthermore, our proposed DiffTalk can be gracefully tailored for higher-resolution synthesis with negligible extra computational cost.

\noindent\textbf{Limitations.} The DiffTalk models talking head generation as an iterative denoising process, which needs more time to synthesize a frame compared with most GAN-based approaches. This is also a common problem of LDM-based works. When driving a portrait with more challenging cross-identity audio, the audio-lip synchronization of the synthesized video is slightly inferior to the ones under self-driven setting. During inference, the network is also sensitive to the mask shape in $z_T$, where the mask needs to cover the mouth region completely and its shape cannot leak any lip shape information. All these inspire our further research directions for superior synthesis results.
Since talking head technology may raise potential misuse issues, we are committed to combating these malicious behaviors and advocate positive applications. Additionally, researchers who want to use our code will be required to get authorization and add watermarks to the generated videos.


\noindent\textbf{Acknowledgment.}
This work was supported in part by the National Key Research and Development Program of China under Grant 2017YFA0700802, in part by the National Natural Science Foundation of China under Grant 62125603, and in part by a grant from the Beijing Academy of Artificial Intelligence (BAAI).

{\small
\bibliographystyle{ieee_fullname}
\bibliography{egbib}
}

\end{document}